\documentclass{article}
\usepackage[accepted]{mlsys2020}

\usepackage{microtype}
\usepackage{graphicx}
\usepackage{subfigure}
\usepackage{booktabs} 
\usepackage{amsfonts}
\usepackage{amsmath}
\usepackage{xspace}
\usepackage{enumitem}
\usepackage{listings}
\usepackage{hyperref}

\usepackage{xcolor}
\newcommand{\latinphrase}[1]{\textit{#1}}
\newcommand{\etal}{\latinphrase{et~al.}\xspace}

\newcommand{\eg}{\latinphrase{e.g.}\xspace}
\makeatletter

\newcommand{\tool}{P2B\xspace}
\newcommand{\MediaMillDifference}{2.6\%\xspace} 
\newcommand{\TextMiningDifference}{3.6\%\xspace} 
\newcommand{\MediaMill}{MediaMill\xspace}
\newcommand{\CriteoDifference}{0.0025\xspace}

\mlsystitlerunning{Privacy-Preserving Bandits}
\usepackage[firstpage]{draftwatermark}
\SetWatermarkText{
 \hspace*{2.4in}
 \raisebox{10in}{
  \includegraphics[height=0.9in]{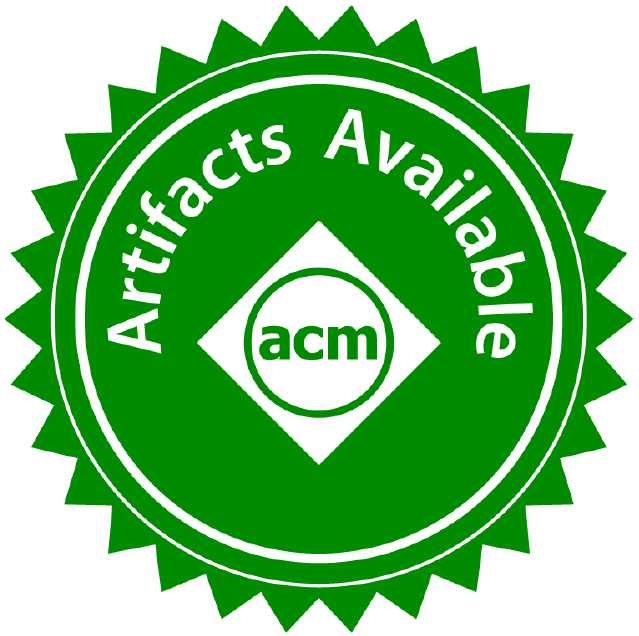}
  \includegraphics[height=0.9in]{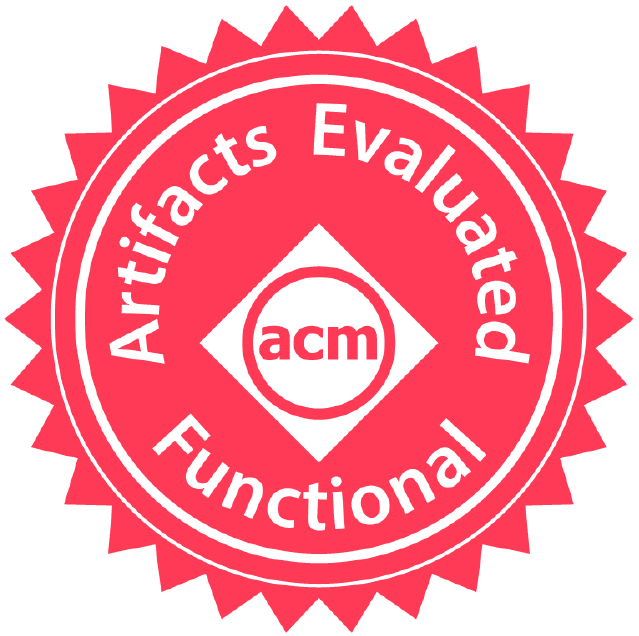}
  \includegraphics[height=0.9in]{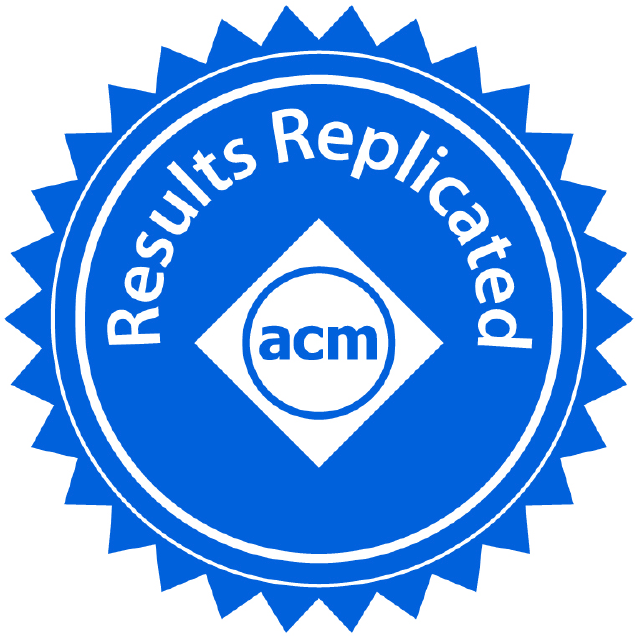}
 }
}
\SetWatermarkAngle{0}
\begin{document}


\twocolumn[
\mlsystitle{Privacy-Preserving Bandits}
\begin{mlsysauthorlist}
\mlsysauthor{Mohammad Malekzadeh}{qmul}
\mlsysauthor{Dimitrios Athanasakis}{brave}
\mlsysauthor{Hamed Haddadi}{brave,icl}
\mlsysauthor{Benjamin Livshits}{brave,icl}
\end{mlsysauthorlist}
\mlsysaffiliation{qmul}{Queen Mary University of London (this work was done during an internship at Brave Software),}
\mlsysaffiliation{brave}{Brave Software,}
\mlsysaffiliation{icl}{Imperial College London}
\mlsyscorrespondingauthor{Mohammad Malekzadeh}{m.malekzadeh@qmul.ac.uk}

\mlsyskeywords{differential privacy, contextual bandit algorithms, recommendation systems, privacy-preserving machine learning} 
\vskip 0.3in
\begin{abstract}
Contextual bandit algorithms~(CBAs) often rely on personal data to provide recommendations. Centralized CBA agents utilize potentially sensitive data from recent interactions to provide personalization to end-users. Keeping the sensitive data locally, by running a local agent on the user's device, protects the user's privacy, however, the agent requires longer to produce useful recommendations, as it does not leverage feedback from other users.

This paper proposes a technique we call \emph{Privacy-Preserving Bandits}~(\tool); a system that updates local agents by collecting feedback from other agents in a differentially-private manner. Comparisons of our proposed approach with a non-private, as well as a fully-private (local) system, show competitive performance on both synthetic benchmarks and real-world data. Specifically, we observed only a decrease of \MediaMillDifference and \TextMiningDifference in multi-label classification accuracy, and a CTR increase of \CriteoDifference in online advertising for a privacy budget $\epsilon \approx 0.693$. These results suggest \tool is an effective approach to challenges arising in on-device privacy-preserving personalization. 
\end{abstract}
]

\printAffiliationsAndNotice{} 

\section{Introduction}

Personalization is the practice of tailoring a service to individual users by leveraging their interests, profile, and content-related information. For example, a personalized news service would learn to recommend news articles based on past articles the user has interacted with~\cite{Li2010}, or a badge achievement system can improve the user engagement by offering gamified activities that are close to the user's interests~\cite{gharibi2017gamified}. Contextual Bandit Algorithms~\cite{Li2010, Ronen2016} are frequently the workhorse of such personalized services. CBAs improve the quality of recommendations by dynamically adapting to users' interests and requirements. The overall goal of the {\em agent} in this setting is to learn a policy that maximizes users engagement through time. In doing so, the agent collects potentially sensitive information about a user's interests and past interactions, a fact that raises privacy concerns. 

On-device recommendation can address privacy concerns, arising from the processing of users' past interactions, by storing and processing any user feedback locally. As no personal data leaves the user's device, this approach naturally maintains a user's privacy\footnote{In a practical setting, the service maintainer still needs to do extra work to eliminate potential side channels attacks.}. However, the on-device approach is detrimental to personalization, as it fails to incorporate useful information gleaned from other users, limiting its utility in making new recommendations and leading to the cold-start problem, where the quality of initial recommendations is often insufficient for real-life deployment.  

Techniques for privacy-preserving data analysis such as the Encode-Shuffle-Analyze~(ESA) scheme implemented in PROCHLO~\cite{Bittau2017, erlingsson2020encode} promise to safeguard user's privacy, while preserving the quality of resulting recommendations, through a combination of cryptographic, trusted hardware, and statistical techniques. In this paper, we explore the question of how the ESA approach can be used in a distributed personalization system to balance the quality of the received recommendations with maintaining the privacy of users.   

We propose \tool---Privacy-Preserving Bandits---a system where individual agents running locally on users' devices are able to contribute useful feedback to other agents through centralized model updates while providing differential privacy guarantees~\cite{Gehrke2012Crowd-blendingPrivacy, Dwork2013}. To achieve this, \tool combines a CBA agent running locally on a user's device with a privacy-preserving data collection scheme similar to ESA. 

Specifically, this paper makes the following contributions: 
\begin{itemize}\itemsep=-1pt
 
\item {\bf Context Encoding}.
We propose simple methods of efficiently encoding feedback instances of a CBA on the user's device. The encoding process combines simple clustering algorithms, such as $k$-means, with the favorable spatial structure of normalized contexts. We study the effects of this structure on both privacy and utility,  and experimentally demonstrate how this encoding approach is competitive with methods that do not take privacy under consideration. 

\item {\bf Privacy Analysis}. 
We examine the overall trade-offs between recommendation quality and privacy loss by performing a differential privacy analysis of our proposal according to the crowd-blending privacy model~\cite{Gehrke2012Crowd-blendingPrivacy}. Our analysis proves that \tool results in a small $\epsilon$ value for differential privacy, which can be directly quantified from the probability of an agent participating in the data collection mechanism. This demonstrates mathematically that \tool provides a concrete and desirable privacy guarantee. 

\item {\bf Experimental Results}. 
We construct a testbed for the evaluation of our approach where our proposal is evaluated on synthetic and real-world data. We include results on a synthetic benchmark, multi-label classification, and online advertising data. Our results experimentally demonstrate that \tool remains competitive in terms of predictive utility with approaches that provide no privacy protections. At the same time, it substantially outperforms on-device cold-start models that do not share data\footnote{Code and data are available at: \url{https://github.com/mmalekzadeh/privacy-preserving-bandits}}.
\end{itemize}
\section{Background}
Contextual bandit algorithms present a principled approach that addresses the exploration-exploitation dilemma in dynamic environments, while utilizing additional, contextual information. In the contextual bandit setting, at time $t'<t$ the agent selects an action $a_{t'} \in \{1, 2, \ldots, |A|\}$ based on the observed  $d$-dimensional context vector  $X_{t'}$ at that time. The agent then obtains the reward $r_{t',a} \in \{0,1\}$ associated with the selected action, without observing the rewards associated with alternative actions. 

Upper confidence bound~(UCB) method computes upper bounds on the plausible rewards of each action and consequently selects the action with the highest bound. In the implemented \tool, we use  {\em LinUCB} \cite{Chu2011ContextualFunctions}, which computes UCB based on a linear  combination of rewards encountered on previous rounds to propose the next action. In LinUCB the exploration-exploitation actions depend on $\alpha\geq 0$, which is the parameter controlling that trade-off.

In the following we provide some preliminary material on privacy preserving data release mechanisms in relation to personalization with contextual bandit algorithms.

\subsection{Differential Privacy Preliminaries}

In the differential privacy framework ~\cite{Dwork2013}  a data sharing mechanism violates its users privacy if data analysis can reveal if a user is included in the data with a degree of confidence higher than the mechanism's bound. 

\textbf{Definition 1: Differentially-Private Data Sharing.}\label{def:dp} Given $\epsilon, \delta \geq 0$, we say a data sharing mechanism $\mathcal{M}$ satisfies $(\epsilon, \delta)-$differential privacy if for all pairs of neighbor datasets of context vectors $\mathbf{X}$ and $\mathbf{X}'$ differing in only one context vector $X$, such that $\mathbf{X} = \mathbf{X'}  \cup \{X\}$,  and for all $R \subset Range(\mathcal{M})$,
\begin{gather*}
Pr[\mathcal{M}(\mathbf{X}) \in R] \leq e^{\epsilon} Pr[\mathcal{M}(\mathbf{X}') \in R] + \delta
\end{gather*}
Intuitively, a differentially-private $\mathcal{M}$ controls the ability of any attacker to distinguish whether or not a specific user's data is used in the computation, holding the participation of all other users fixed~\cite{kifer2020guidelines}.

\begin{table}[t]
\caption{Notation used in this paper.}\label{tab:notation}
\begin{center}
\begin{small}
\begin{tabular}{p{1.8cm} p{5.7cm}} 
\toprule
$A$ & set of available actions;\\
$a_{t'}\in A$ & chosen action at time $t'<t$;\\
$d\in \mathbb{N}$ & size of the context vector; \\
$k \in \mathbb{N}$ & number of possible {\em encoded} contexts.\\
$n \in \mathbb{N}$ & number of possible {\em original} contexts;\\
$l\in \mathbb{N}$ & crowd-blending threshold; \\
$p \in [0,1]$ & probability of pre-sampling;\\ 
$q \in \mathbb{N}$ & precision of digits after the decimal point;\\
$r_{t',a} \in \{0,1\}$ & reward associated to an action $a_{t'}$;\\
$t\in \mathbb{N}$ & number of interactions with user;\\ 
$u\in \mathbb{N}$ & number of users;\\ 
$X_{t'}\in \mathbb{R}^{d}$ & $d$-dimensional context vector at time $t'<t$;\\ 
$\mathbf{X}$ & dataset of context vectors;\\
${y}_{t'}\in \mathbb{N}$ & encoded context vector associated to an $X_{t'}$;\\
$z \in \mathbb{R}$  & random Gaussian noise; \\ 
$\alpha \in \mathbb{R}^{+}$ &  exploration-exploitation ratio in LinUCB;\\
$\epsilon, \delta \in \mathbb{R}^{+}$ &  parameters of  differentially privacy.\\ 
\bottomrule
\end{tabular}
\end{small}
\end{center}
\vskip -0.1in
\end{table}

\subsection{Crowd-blending}
Crowd-blending privacy model ~\cite{Gehrke2012Crowd-blendingPrivacy} is a relaxation of differential privacy. In the crowd-blending framework a user blends with a number $l \geq 1$ of users in a way that replacing the user's data with any other individual in this crowd does not alter the results of statistical queries. A necessary condition for this approach is the existence of $l-1$ other users that blend with the user. If the data does not contain a sufficient number of other individuals the mechanism essentially ignores the user's data. Consequently an {\em encoding} mechanism $\mathcal{M}(\cdot)$ that satisfies crowd-blending privacy~\cite{Gehrke2012Crowd-blendingPrivacy} can be formalized as follows: 

\textbf{Definition 2: Crowd-Blending Encoding.} Given $l\geq1$, we say an encoding mechanism $\mathcal{M}$ satisfies \mbox{$(l, \bar{\epsilon}=0)-$}crowd-blending privacy if for every context vector $\mathbf{x}$ and for every dataset of context vectors $\mathbf{X} = \mathbf{X'}  \cup \{X\}$ we have 
\begin{gather*}
\Big\vert\big\{y \in \mathcal{M}\big(\mathbf{X}\big) :  y=\mathcal{M}\big(\{X\}\big)\big\}\Big\vert \geq l
\enspace\text{or}\enspace
 \mathcal{M}(\mathsf{\mathbf{X}}) =  \mathcal{M}(\mathsf{\mathbf{X}'}), 
\end{gather*}
where $\big\vert\{\cdot\}\big\vert$ denotes the size of a set\footnote{Note that $\mathbf{X'}$ can be any context dataset not including $X$.}.

This means that for every context vector $X_{t'}$, either its encoded value $y_{t'}$ blends in a crowd of $l-1$ other values in the released dataset, or the mechanism does not include $X_{t'}$ at all. In this setting, $\bar{\epsilon} = 0$ means that if $\mathcal{M}(\cdot)$ releases an encoded value $y_{t'}$ for a given $X_{t'}$, it should be exactly the same as other $\l-1$ encoded values coming from other $\l-1$ context vectors. Every $(\epsilon, \delta)$-differentially private mechanism also satisfies $(\mathrm{l}, \epsilon, \delta)$-crowd-blending privacy for every integer $\l \geq 1$~\cite{Gehrke2012Crowd-blendingPrivacy}. 

An important aspect of crowd-blending privacy is that if a random {\em pre-sampling} is used during data collection, before applying the crowd-blending on the collected data, the combined pipeline provides zero-knowledge~\cite{Gehrke2011TowardsPrivacy} privacy and thus differential privacy~\cite{Dwork2013} as well. This intuitively says that if my data blends in a crowd of $l$ other people, then other people similar to me can easily take the place of me. Therefore, my data being sampled or not has a negligible effect on the final output of the combined mechanism. \tool utilizes this property. 

\subsection{Differentially-private Data Collection}

Privacy-preserving data collection has numerous applications. To collect browser settings data, Google has built RAPPOR into the Chrome browser~\cite{Erlingsson2014}. RAPPOR hashes textual data into Bloom filters using a specific hash function. It then randomizes the Bloom filter, that is a binary vector, and uses it as the permanent data to generate instantaneous randomized response for sharing with the server. RAPPOR can be used to collect some aggregated statistics from all users. For example, having a huge number of users, it can estimate the most frequent homepage URLs. However, since the shared data is only useful for estimating aggregated statistics like mean or frequency, the utility of every single shared data for training a model is often too low~---~the accuracy of the private data becomes unacceptable even with a large number of users.

As an extension to RAPPOR, the ESA architecture~\cite{Bittau2017, erlingsson2020encode} adds two more layers to the LDP layer, called shuffler and analyzer, that are aimed to obscure the identity of the data owner  through oblivious shuffling with trusted hardware. The Shuffler in the ESA architecture eliminates all metadata attached to the users' reports to eliminate the possibility of linking a data report to a single user during data collection. However, if users do not want to trust any other party, the provided privacy will be the same as RAPPOR. 
\section{Methodology}

In \tool, every user runs their own CBA agent that works independently of any other agents\footnote{Note that {\em user} is an individual and {\em agent} is an on-device algorithm that serves a user.}. At time $t'$ the agent observes the current context $X_{t'}$, which represents recent user's activities (\eg recent web browsing history). Based on $X_{t'}$, the agent proposes an action $a_{t'}$, and consequently observes the reward $r_{t',a}$ associated with the action (\eg proposing a news article and observing the user's reaction to that article). As the interaction proceeds locally, we refer to agents running on a user's device as {\em local agents}. The agent may periodically elect to send some information about an observed interaction to a data collection server. Using the collected data, the server updates a central model that is then propagated back to the users.  

\begin{figure*}[ht]
    \centering
    \includegraphics[width=\textwidth]{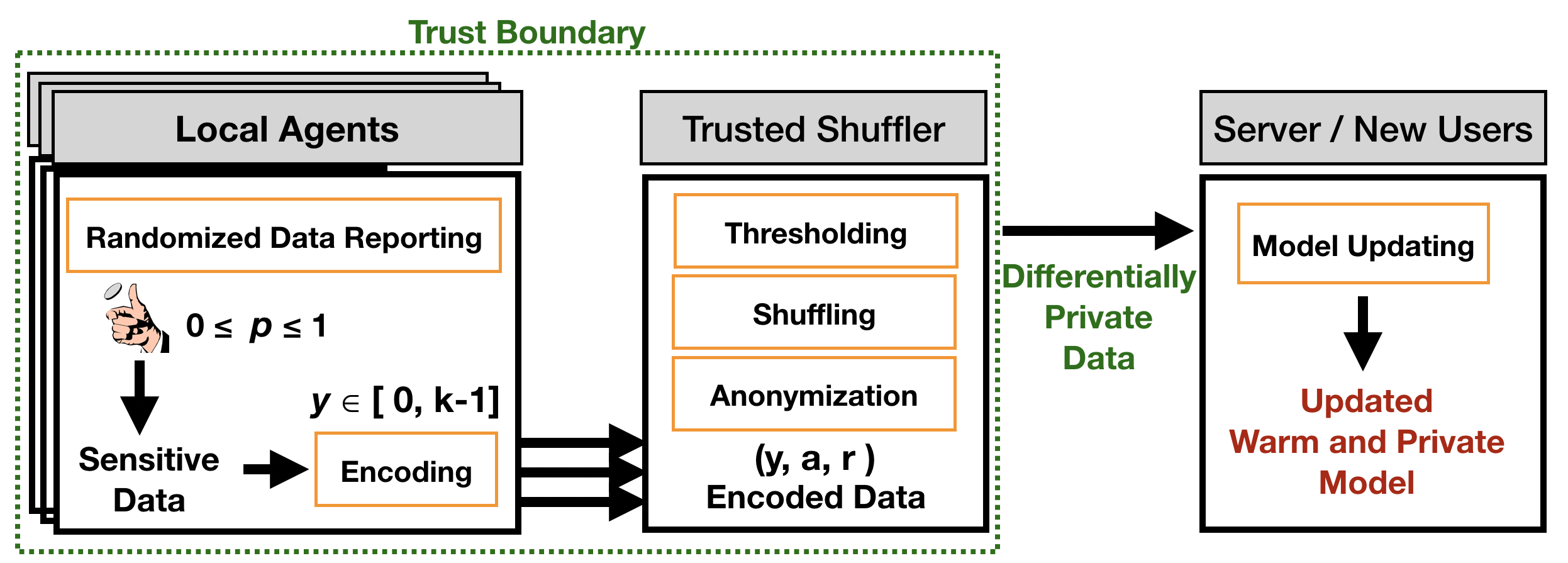}
    \caption{System architecture for \tool.}
    \label{fig:arch}
\end{figure*}

Figure~\ref{fig:arch} summarizes the overall architecture of the proposed framework. The rest of this section presents \tool's operation and details its various components. Section~\ref{sec:privacy} describes how \tool satisfies differential privacy, and Section~\ref{sec:eval} experimentally demonstrates how this approach improves the performance of local agents. 

\subsection{Randomized Data Reporting}
\label{sec:sampling}

Local agents participate in \tool through a randomized participation mechanism. After some interactions with the user, $t \geq 1$, the local agent may randomly construct a payload, containing an encoded instance of interaction data, with probability $p$. Randomized participation is a crucial step in two ways. First, it raises the difficulty of re-identification attacks by randomizing the timing of recorded interactions. More importantly, the participation probability $p$ as a source of randomness, has a direct effect on the differential privacy parameters $\epsilon$ and $\delta$ we will establish in Section~\ref{sec:privacy}. Briefly, by choosing the appropriate participation probability, one can achieve any level of desired privacy guarantee in \tool.

\subsection{Encoding}
\label{sec:encoding}

The agent encodes an instance of the context prior to data transmission. The encoding step acts as a function that maps a $d$-dimensional context vector $X$ into a code ${y} \in \{0,1,\ldots,k-1\}$, where $k$ is the total number of encoded contexts. Agents encode a normalized context vector of fixed precision that uses $q$ digits for each entry in the vector. An example of this representation approach is a normalized histogram, where entries in the histogram sum to~$1$ and are represented to a precision of $q$ decimal digits. This combination of normalization and finite precision has two important characteristics.
 
First, it is possible to precisely enumerate all possible contexts according to the {\em stars and bars} pattern in combinatorics~\cite{benjamin2003proofs}. Specifically the cardinality~$n$ of the set of normalized context vectors $X \in \mathbf{X}$, using a finite precision of~$q$ decimal digits is  
\begin{equation}\label{eq:s_and_b}
n = \binom{10^q+d-1}{d-1}.
\end{equation}
Secondly, sample points in the normalized vector space are distributed uniformly in a grid shape. Given that the agent tends to propose similar actions for similar contexts, neighboring context vectors can be encoded into the same context  code $y$.  While this approach may appear to be limiting, it generalizes to other instances where the context vectors are bounded and exhibit some degree of clustered structure. 

\begin{figure}[tb]
    \centering
    \includegraphics[width=\columnwidth]{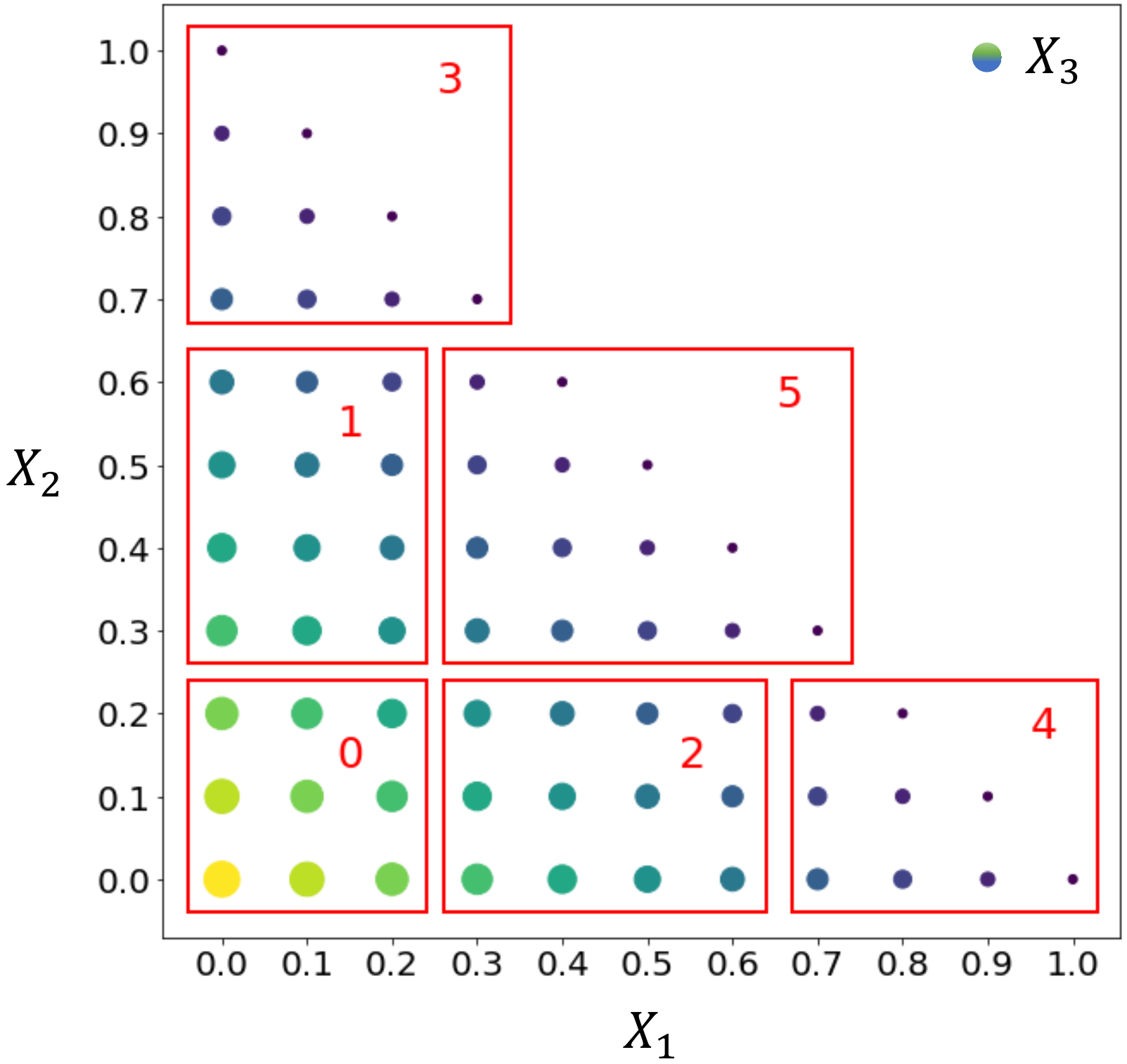}
    \caption{Normalized vector space of $X=(X_1,X_2, X_3)$ with $q=1$ and cardinality~$n=66$. In this setting, we have normalized vectors such that $ X_1 + X_2 + X_3 = 1$. Thus, size of the circles shows the value of $X_3$. Rectangles show a potential encoding of the vector space with $k=6$ and minimum cluster size $l=9$.}
    \label{fig:dim3space}
\end{figure}

Figure \ref{fig:dim3space}, provides a concrete example of the encoding process on a $3$-dimensional vector space being encoded in $k=6$ different codes.  The cardinality of the encoding, $n$, is important in establishing a desired utility-privacy trade off. The encoding algorithm can be chosen depending on application requirements, but we limited our experimental evaluation to the simple $k$-means clustering~\cite{Sculley2010Web-scaleClustering}. 

After electing to participate and encoding an instance of a user interaction, the agent transmits the data in the form of the tuple $(y_{t'}, a_{t'}, r_{t',a})$ to the {\em shuffler}. 

\subsection{Shuffler}
The trusted shuffler is a critical part of every ESA architecture~\cite{Bittau2017, erlingsson2020encode}, and in \tool it is necessary for ensuring anonymization and crowd-blending. Following the same PROCHLO implementation~\cite{Bittau2017}, the shuffler operates in a secure enclave on trusted hardware and performs three tasks:
\begin{enumerate}
    \item \textbf{Anonymization}: eliminating all the received meta-data (\eg IP address) originating from local agents. 
    \item \textbf{Shuffling}: gathering tuples  received from several local agents into batches and shuffling their order. This helps to prevent side-channel attacks (\eg timing attack).
    \item \textbf{Thresholding}: removing tuples that their encoded context vector frequency in the batch is less than a defined {\em threshold}. This threshold plays an important role in the achieved differential privacy bound.
\end{enumerate}

After performing these three operations, the shuffler sends the refined batch to the {\em server} for updating the model. Upon receiving the new batch of training data, the server updates the global model and distributes it to local agents that request it. More importantly, this global model is used as a warm-start by new local agents that are joining to the system to help in preventing the cold-start problem.
\section{Privacy Analysis}\label{sec:privacy}

This section analyzes how \tool ensures differential privacy through a combination of pre-sampling and crowd-blending. As context vectors are multi-dimensional real-valued vectors, we assume a context vector in its original form can be uniquely assigned to a specific user. \tool assumes no prior on the possible values for a context vector, meaning context vectors are coming from  a uniform distribution over the underlying vector space. With these assumptions and the provided zero-knowledge privacy, \tool can resist strong adversaries with any kind of prior knowledge or side information.

For the $n$ different context vectors in Equation~(\ref{eq:s_and_b}), the optimal encoder (Section~\ref{sec:encoding}) encodes every $n/k$ contexts into one of the possible $k$ codes. Consequently, when a total number of $u$ users participate in \tool to sends a tuple to the server, the optimal encoder satisfies crowd-blending privacy with $l=u/k$.  In the case of a suboptimal encoder, we consider $l$ as the size of the smallest cluster in the vector space. Furthermore, situations where the number of users is small, leading to a small $l$, can be addressed by adjusting the shufflers {\em threshold} to reach the desired $l$. Essentially, $l$ can always be matched to the shuffler's threshold.

Each user randomly participates in data sharing with probability $p$ (Section~\ref{sec:sampling}), and then encoding the pre-sampled data tuple. Following \cite{Gehrke2012Crowd-blendingPrivacy}, the combination of (i) pre-sampling with probability $p$ and (ii) $(l,  \bar{\epsilon})-$crowd-blending, leads to a differentially private mechanism with
\begin{equation}\label{eq:eps}
\epsilon=\ln\Big(p\cdot\big(\frac{2-p}{1-p}\cdot e^{\bar{\epsilon}}\big)+\big(1-p\big)\Big)
\end{equation}
and
\begin{equation}\label{eq:delta}
\delta=e^{-\Omega(l\cdot(1-p)^2)}.
\end{equation}

Here $\Omega$ is a constant that can be calculated based on the analysis provided by \cite{Gehrke2012Crowd-blendingPrivacy}. 

\begin{figure}[bt]
    \centering
    \includegraphics[width=\columnwidth]{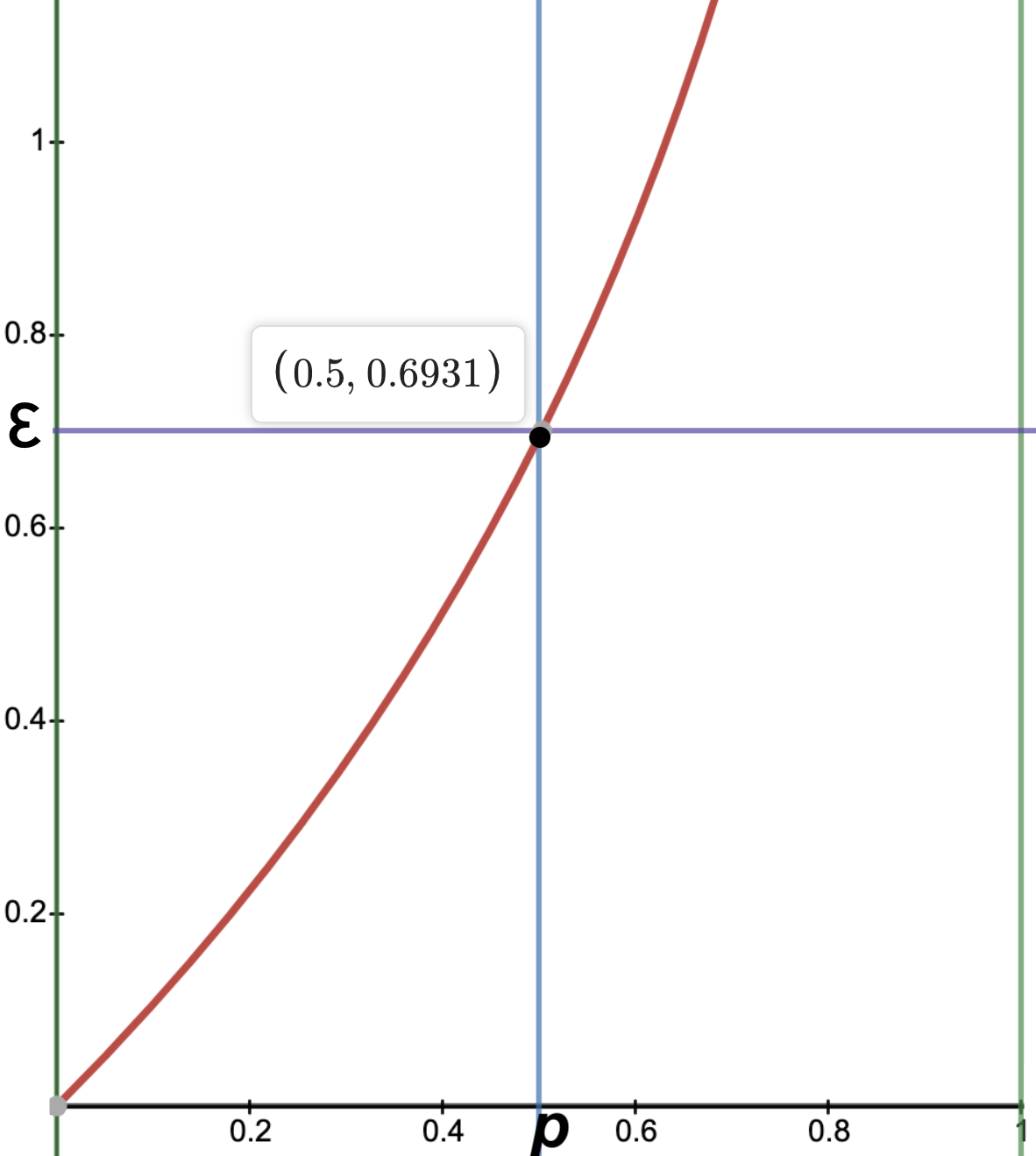}
    \caption{$\epsilon$ as a result of the probability of participating in the scheme $p$. }
    \label{fig:epsilonProb}
\end{figure}

Our encoding scheme provides an $\bar{\epsilon} = 0$ for crowd-blending, as the encoded values for all the members of a crowd is exactly the same. As a consequence, the $\epsilon$ parameter of the differential privacy of the entire data sharing mechanism depends entirely on the probability $p$ of participation as (Figure \ref{fig:epsilonProb}): 

\begin{equation*}
\epsilon = \ln\Big(p\cdot\big(\frac{2-p}{1-p}\big)+\big(1-p\big)\Big).
\end{equation*}

For example by trading half of the potential data ($p=0.5$) \tool achieves an $\epsilon \approx~0.693$ that is a strong privacy guarantee. On the other hand, $\delta$ depends on both $p$ and $l$. To understand the effect of $\delta$, Dwork~\etal~\cite{Dwork2013} prove that an $(\epsilon,\delta)$-differentially private mechanism ensures that for all neighboring datasets, the absolute value of the privacy loss will be bounded by $\epsilon$ with probability at least $1-\delta$. Therefore, by linearly increasing the crowd-blending parameter $l$, we can exponentially reduce the $\delta$ parameter. 

Our illustrations here provide concrete settings and parameters that show how \tool satisfies the requirements of the differential privacy model proposed in~\cite{Gehrke2012Crowd-blendingPrivacy}. We refer the interested reader to the rigorous analysis and proofs provided in the original paper~\cite{Gehrke2012Crowd-blendingPrivacy}. 
\section{Exprimental Evaluation}
\label{sec:eval}

This section explores \tool's privacy-utility trade-offs compared to (i) competing non-private and (ii) completely private approaches. The experimental evaluation assumes the standard bandit settings, where the local agent learns a policy based on the contextual Linear Upper Confidence Bound algorithm (LinUCB)~\cite{Chu2011ContextualFunctions, Li2010}. For simplicity, throughout the experiments, the probability $p$ of randomized transmission by a local agent was set to $p=0.5$, the rounding parameter $q$ of the encoder was set to $q=1$, and the parameter $\alpha$ for LinUCB was set to $\alpha=1$, meaning that the local agent is equally likely to propose an exploration or exploitation action. The experiments compare the performance of the following three settings:

{\bf Cold.} The local agent learns a policy without any communication to the server at any point. As there is no communication, this provides full privacy, but each agent has to learn a policy from a cold-start. 
  
{\bf Warm and Non-Private.} In this setting, local agents communicate the observed context to the server in its original form. Thus, other agents are able to initialize their policy with a model received from the server and start adapting it to their local environment. This is called {warm and non-private} start, and represents the other end of the privacy/utility spectrum with no privacy  guarantee afforded to the users.

{\bf Warm and Private} In this setting, local agents communicate with the server using \tool. Once more, other agents initialize their internal policy with an updated model received from the server and start to adapt it to their local environment. We term this a {warm and private} start, and the provided privacy guarantees function according to the analysis in section \ref{sec:privacy}. 

These approaches are evaluated on synthetic benchmarks, two multi-label classification datasets, and an online advertising dataset. 

\subsection{Synthetic Preference Benchmark} \label{subsec:syn}

These benchmarks consider the setting where there's a stochastic function $\mathcal{F}$ that relates context vectors with the probability of a proposed action receiving a reward. Specifically, $\mathcal{F}$ is the scaled softmax output of a matrix-vector product of the user contexts with a randomly generated weight matrix $W$: 

\begin{equation}\label{eq:synth}
\mathcal{F}(X) = \mathsf{softmax}(W\cdot X+b) \in [0,1]^{|A|}.
\end{equation}
For the specific implementation in this section, we use the default weight initializer of TensorFlow~\cite{abadi2016tensorflow} to set $W$ and $b$, and then keep them fixed through the experiment.
In the LinUCB algorithm, we set the mean reward $\bar{r}_{a}$ for every action $a_t$ as
\begin{equation*}
\bar{r}_{a} = \beta \mathcal{F}_a(X)+z,
\end{equation*}
where $\mathcal{F}_a$ is the $a$-th component of the softmax output in Equation~\ref{eq:synth}, $\beta$ is a scaling factor with $0\le \beta \leq 1$, and $z$ is a random Gaussian noise $z\sim\mathcal{N}(0,\sigma^2)$. Hence, to create an instance of a synthetic interaction between user and agent, we first  randomly and uniformly generate a context vector $\mathbf{x}$. Then having $\mathbf{x}$ and using $\mathcal{F}$, for each possible action $a$ we calculate the probable reward $r$. For time $t'<t$, this process results to a synthetic data tuple $(X_{t'}, a_{t'}, r_{t',a})$.

For all synthetic benchmarks the parameters are set as follows. The scaling factor for the preference function was fixed to $\beta =0.1$ and the variance of the Gaussian noise $\sigma^2=0.01$. In terms of local agent settings, the agents use $k=2^{10}$ codes for encoding purposes and observe $t=10$ local interactions before randomly transmitting an instance with probability  $p=0.5$. We varied the number of dimensions $d$ of the context vector in the range of $6$ to $20$, and the number of actions in the range of $10$ to $50$. We observe the average reward in each setting as the user population $u$ grows from $10^3$ to $10^6$.

\begin{figure}[tb]
    \centering
	\begin{minipage}[t]{\linewidth}
		\includegraphics[width=\linewidth]{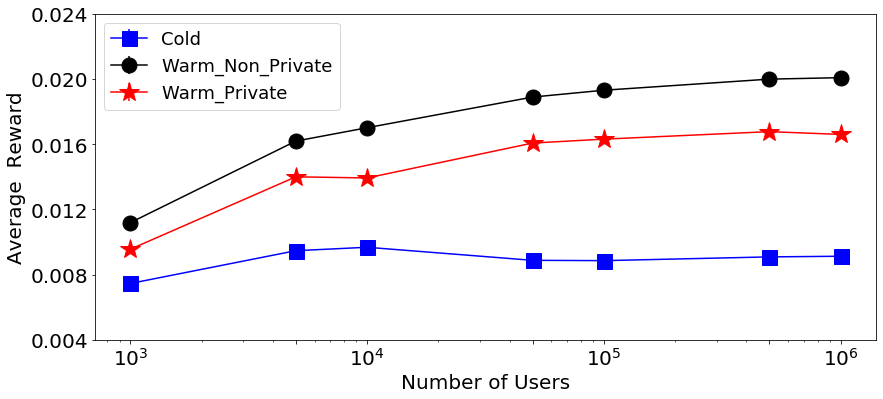}
	\end{minipage}
	\vfill
	\begin{minipage}[t]{\linewidth}
		\includegraphics[width=\linewidth]{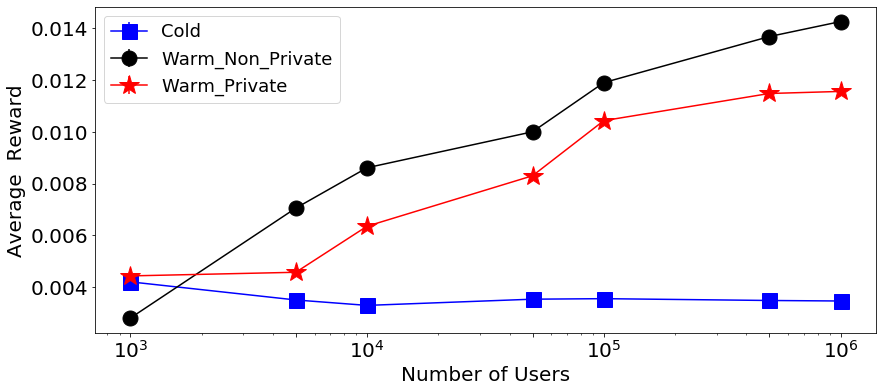}
	\end{minipage}
	\vfill
	\begin{minipage}[t]{\linewidth}
		\includegraphics[width=\linewidth]{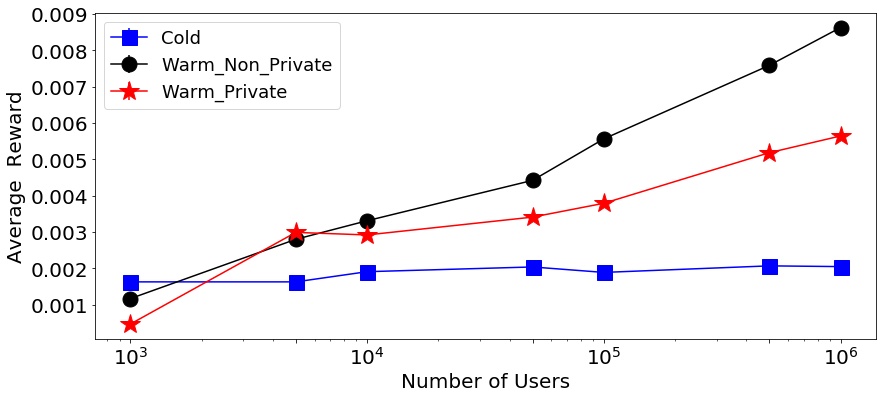}
	\end{minipage} 
 \caption{Synthetic benchmarks: (Top) $|A|=10$, (Middle) $|A|=20$ (Bottom) $|A|=50$. For all: $d=10$ and $t=10$. The expected reward in this setting has a strong dependence on the number of actions as agents will spend considerable time exploring alternative actions. }
    \label{fig:syn_1}
\end{figure}

Our results in Figure~\ref{fig:syn_1} indicate that for a small number of interactions the cold-start local model fails to learn a useful policy. By contrast, the warm models substantially improve as more user data becomes available. Utilizing prior interaction data in this setting, more than doubles the effectiveness of the learned policies, even for relatively small user populations. Overall, the non-private agents have a performance advantage, with the private version trailing. 

\begin{figure}[tb]
    \centering
    \includegraphics[width=\columnwidth]{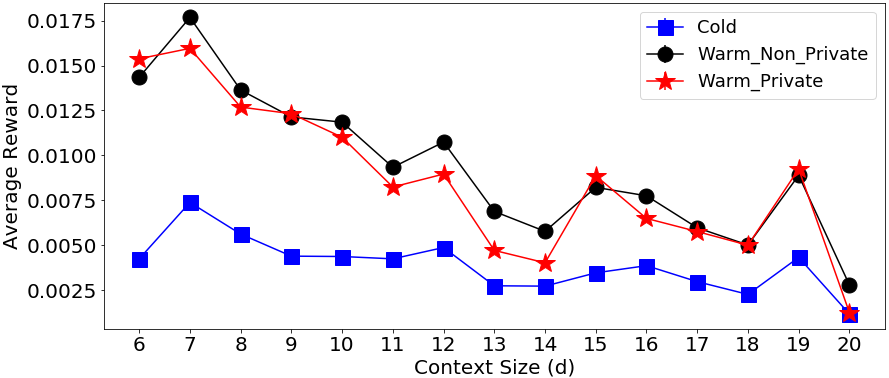}
    \caption{Synthetic benchmarks: $u=20000$, $|A|=20$, $t=20$, and $d =\{6,7,\ldots,20\}$. As the dimensionality of the context increases the average reward for this settings is reduced as agents spend more time trying to explore their environment. }  
    \label{fig:d_range}
\end{figure}

Figure~\ref{fig:d_range} illustrates how the dimensionality of the context vector affects the agent's expected reward. By increasing $d$ from 6 to 20 the average reward decreases as agents spend more time exploring the larger context space. \tool remains competitive with its non-private counterparts, and on occasion outperforms them, especially for low-dimensional context settings. The number of actions has a similar effect to the dimensionality of the context, as agents have to spend more time exploring suboptimal actions. Once more, the results  of Figure \ref{fig:syn_1} indicate that this is the case experimentally. 

The number of users is pivotal in the trade-off between privacy and utility. For simplicity, and in order to avoid unnecessary computation, experiments fix $p = 0.5$. However, the expected reward for other $p$ values can be simply deduced from the current experiments. For example, if $p$ changes from $0.5$ to $0.25$ (meaning $\epsilon$ is changed from $0.69$ to $0.28$) \tool would require $100K$ users rather than $50K$ to achieve the same performance with the new privacy guarantee. In simple terms, to achieve smaller $\epsilon$, a smaller $p$ is required, which consequently means  a proportionally increased number of users is necessary to maintain the same performance.

\begin{figure}[t]
    \centering
	\begin{minipage}[t]{\columnwidth}
		\includegraphics[width=\linewidth]{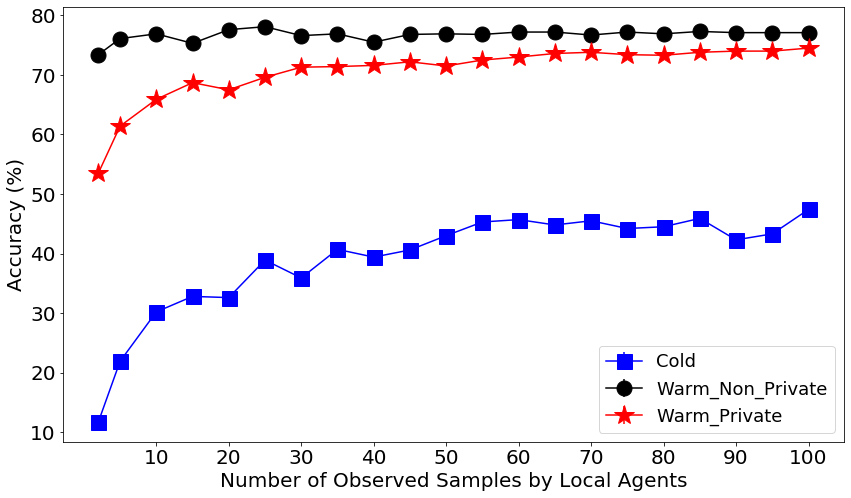}
	\end{minipage}
	\vfill
	\begin{minipage}[t]{\columnwidth}
		\includegraphics[width=\linewidth]{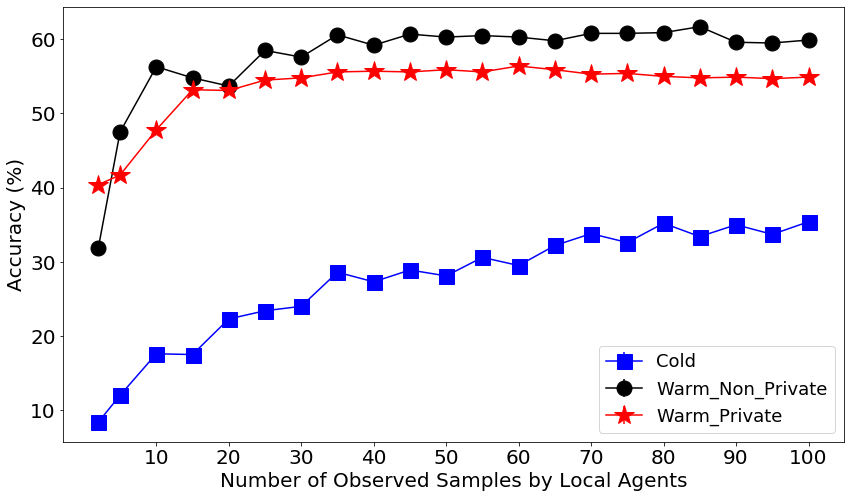}
	\end{minipage} 
 \caption{Accuracy in multi-label datasets: (Top) Media-Mill with $d=20$ and $|A|=40$ (Bottom) Text-Mining with $d=20$ and $|A|=20$. As local agents observe more interactions they obtain better accuracy. This has a multiplicative effect in the distributed settings where agents reach to the plateau much faster.}
    \label{fig:mlc}
\end{figure}


\subsection{Multi-Label Classification}\label{subsec:mlc}   

These experiments examine the performance of the three settings on multi-label classification with bandit feedback. We consider two datasets, namely 
(1) MediaMill~\cite{Snoek2006TheMultimedia}, a video classification dataset including~43,907 instances,~120 extracted features from each video, and 101 possible categories, and 
(2) a TextMining dataset~\cite{Srivastava2005DiscoveringSystems} including~28,596 instances,~500 extracted features from each text, and~22 possible categories.

\tool's performance in this setting was evaluated according to the following setting. We consider a fixed number of local agents. Each agent has access to, and is able to interact with a small fraction of the dataset. In particular every agent has access to up to $100$ samples, which were randomly selected without replacement from the entire dataset. 70\% of agents to participate in \tool and we test the accuracy of the resulting models with the remaining~30\%. For both datasets, the local agents use $k=2^{5}$ codes for encoding purposes. 

This setting is particularly interesting as it allows us to study how the predictive performance of the system changes when the local agents interact more with the user. In terms of relative performance between the different approaches, the results in Figure~\ref{fig:mlc} repeat the findings of the synthetic benchmarks. The non-private warm version is better than the private warm version, which is still better than the cold version that utilizes only local feedback. 

However, we can also observe that the cold version given enough interactions produces increasingly improving results. The centralized update mechanism tends to have a multiplicative effect, especially when there is little local interaction data before reaching a plateau. 

\subsection{Online Advertising}  \label{subsec:oad}
Here, we consider an advertisement recommendation scenario where the action is to recommend an ad from one of the existing categories. We use the Criteo dataset from a Kaggle contest\footnote{\url{https://labs.criteo.com/category/dataset}}. It consists of a portion of Criteo's traffic over a period of~7 days, including~13 numerical features and~26 categorical features. For each record in the dataset there is a label that shows whether the user has clicked on the recommended ad or not. For anonymization purposes, the feature's semantics are unknown and the data publisher has hashed the values of categorical features into~32 bits. 

As the exact semantics of the features were not disclosed, we assume that numerical features represent user's context and categorical features correspond to the type of the proposed product. For each sample in the dataset, we hash the values of the 26 categorical features into an integer value. The resulting integer value is then used as possible product category to recommend. The hashing procedure operates as follows: First, the ~26 categorical values are reduced into a single hashed value using feature hashing~\cite{Weinberger2009FeatureLearning}. After hashing, the~40 most frequent hash codes are selected. These are converted into an integer value in the range between ~1 to~40, based on their frequency (label~1 shows the most frequent code and so on). Finally, for evaluation we only use data samples having one of this 40 values as the product label and ignore the remaining data.

\begin{figure}[t]
    \centering
	\begin{minipage}[t]{\columnwidth}
		\includegraphics[width=\linewidth]{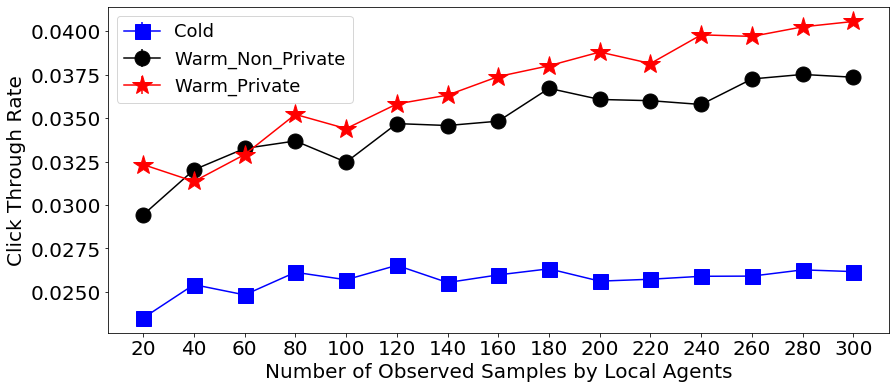}
	\end{minipage}
	\vfill
	\begin{minipage}[t]{\columnwidth}
		\includegraphics[width=\linewidth]{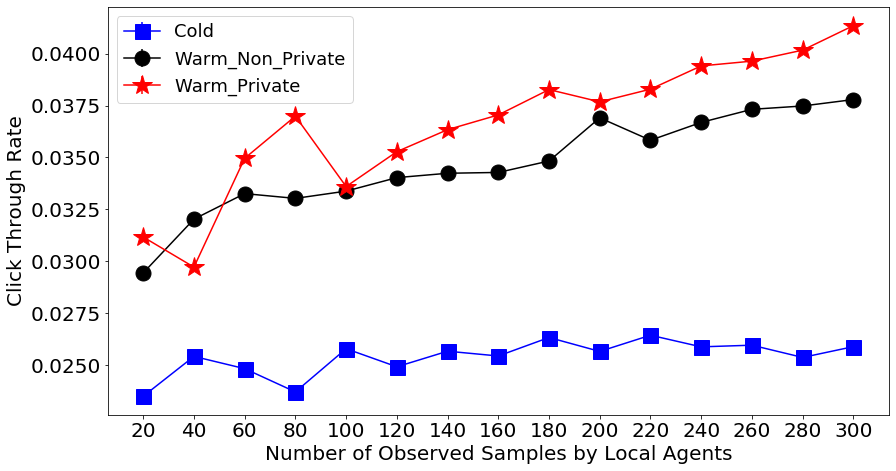}
	\end{minipage} 
 \caption{Criteo results. $d=10$, $|A|=40$, (Top) $k=2^5$, and (Bottom) $k=2^7$. The private and non-private agents obtain similar performances for low numbers of local interactions. As the number of local interactions increase the private agents perform better than their non-private counterparts. }
    \label{fig:criteo}
\end{figure}

During evaluation, the local agent observes the values of the numerical features as the context vector, and in response takes one of the $|A|=40$ possible actions. The agent obtains a reward of 1, if the proposed action matches the logged action in the dataset.  The remaining experimental setup for the Criteo dataset is similar to the one used in the multi-label datasets. Specifically, we present experimental results (Figure~\ref{fig:criteo}) for  $d$=10 and total number of actions $|A|=40$. We compare the results for two values of encoding parameter $k=2^5$ and $k=2^7$. The sampling probability $p$ remains $0.5$, and the shuffling threshold remains~10. This experimental setting has $u=3000$ agents, and each of the agents accumulates~$300$ interactions. 

The results in this setting are quite surprising as private agents attain a better click-through rate than their non-private counterparts. There is a recent work~\cite{Nasr2018MachineRegularization} showing that privacy-preserving training of machine learning models can aid generalization as well as protect privacy. There are some factors we believe help explain these experimental results. 

Private agents use the encoded value as the context. As a result, the number of possible contexts is much smaller compared to using the original $d$-dimensional context vector. As contextual bandits need to balance exploration with exploitation, especially in the early stages, a bandit with a smaller context size can quickly reach better results. Furthermore, \tool clusters similar context vectors into the same categories and this also helps a private bandit to better act in similar situations. This is an effect that also can be seen for the higher dimensional contexts of the synthetic benchmarks in Figure \ref{fig:d_range}.

Once more, the number of users plays a critical role in establishing an acceptable trade-off between utility and privacy. As Section~\ref{sec:privacy} explains, $\delta$ in Equation~\ref{eq:delta} can be interpreted as the probability that the worst-case happens in differentially private data sharing. Using a larger $k$ in a system with a small number of users $u$, limits the crowd blending parameter $l$ and consequently leads to undesired potential values for $\delta$. Unfortunately, the number of users for real-world datasets used in this paper is limited posing constraints on the range of parameters in the experimental evaluation.

\section{Related Work}

User data fuels a variety of useful machine learning applications, but at the same time comes with a number of ethical and legal considerations. The quest to balance the seemingly conflicting concerns of privacy and utility have shaped the resurgence of distributed machine learning approaches that safeguard the privacy of user data. This section compares \tool to some recent proposals in this area.

On-device inference ensures that data does not need to leave the device for inference, which is an important step towards protecting privacy. Federated learning techniques extend this approach to doing on device training through the idea of ``bringing the code to the data, instead of the data to the code''. \cite{Bonawitz2019TowardsDesign} proposes and evaluates a high-level system to perform  federated learning across devices with varying availability. Their proposal relies on secure aggregation~\cite{Bonawitz2017, McMahan2017FederatedData} to ensure that user's data remain encrypted even in the server's memory. 

Compared to \tool, federated learning is substantially more complicated in terms of orchestration. Relying on a local model training procedure, it puts some overhead on the user's device and there is substantial communication complexity involved in communicating model updates and training batching. By contrast, \tool puts almost no overhead on user's devices as they only need to run a distance preserving encoding algorithm~\cite{Sculley2010Web-scaleClustering, Aghasaryan2013OnPersonalization} that has the $O(kd)$ complexity at the inference time, and any data exchange happens opportunistically, without requiring any synchronization between clients. 

Moreover, the scope of \tool is personalization. This means \tool is concerned with developing individual models adapted to individual users' preferences, whereas federated learning is primarily concerned with improving a global model.  \tool addresses a personalization problem which is closer to the reinforcement learning setting, than supervised learning. In our setting, feedback is partial, and there can be more than one valid responses for the exact same context. This makes a direct utility comparison with supervised learning methods such as federated learning~\cite{McMahan2017FederatedData}, or differentially private deep learning~\cite{abadi2016deep} infeasible. 

\tool, however, provides an interesting performance in terms of complexity. It leverages existing PROCHLO infrastructure~\cite{Bittau2017} without needing modifications to the aggregate privatization layer. As illustrated in Figure~\ref{fig:arch}, the total scope of changes required for \tool to work is only a clustering lookup table and procedure on the client, allowing P2B to use an existing ESA deployment without modification, and the central server only needs to do simple contextual bandit updates.

The ``Draw and Discard'' machine learning system described in  \cite{Pihur2018Differentially-PrivateLearning} is an interesting combination of the approach of bringing the code to the data with differential privacy guarantees. Their proposed system encompasses decentralized model training for generalized linear models. It maintains $k$ instances of the model at the server side. The draw and discard approach consists of two components: On the server side, it sends a randomly chosen model instance to the client.  Upon receiving an updated model instance, the server randomly replaces it with one of the $k$ existing instances. On the user side, users update the linear model on their data, and add Laplacian noise to the model parameters to satisfy differential privacy. 

Privacy analysis of ``Draw and Discard'' is in the feature-level that assumes features of the model are all uncorrelated. As a consequence, for achieving model-level privacy, where most features are correlated to each other, substantially more noise needs to be added to the updated model before sending it to the server. \tool  does not make any assumption on the correlation among context vector features and it provides a constant and similar privacy guarantee across all the users. Our experiments only consider situations where users participate by sending only one data tuple. However, in the case of collecting $m$ data tuple from each user, because of the composition property of the provided differential privacy~\cite{Dwork2013}, \tool still guarantees ($m\epsilon$)-differential privacy. 

Shariff~\etal~\cite{Shariff2018DifferentiallyBandits} discuss that adhering to the standard notion of differential privacy needs to ignore the context and thus incur linear regret. Thus, they use a relaxed notion of DP, called ``joint differential privacy'', that allows to use the non-privatized data at time $t'$ , while  guaranteeing that all interactions with all other users at timepoints $t''>t'$ have very limited dependence on that user data at time $t'$. Our approach guarantees differential privacy to all users, independent of time of participation in \tool. 

Basu~\etal~\cite{Basu2019DifferentialCost} argue that if a centralized bandit algorithm only observes a privatized version of the sequence of user's actions and responses, based on $\epsilon$-LDP, the algorithm's {\em regret} scales as a multiplicative factor $\epsilon$. Other LDP mechanisms proposed for bandits algorithms~\cite{Tossou2017, Gajane2018CorruptPrivacy} consider a situations where only responses are considered to be private without incorporating contextual information in the learning process.

Although individual components of \tool, namely contextual bandit algorithms and differentially private mechanisms for data collection, have been widely studied, what \tool achieves is combining these components in a novel way that allows controlling the trade-off between overall utility and privacy, relying solely on two straightforward parameters: (i) the probability of participation in the randomized data collection, and (ii) the chosen threshold for the crowd-blending.
\section{Conclusions}
This paper presents \tool, a privacy-preserving approach for machine learning with contextual bandits. We show experimentally that standalone agents trained on an individual's data require a substantial amount of time to learn a useful recommendation policy. Our experiments show that sharing data between agents can significantly reduce the number of required local interactions in order to reach a useful local policy. 

We introduced a simple distributed system for updating agents based on user interactions. In a series of experiments, \tool shows considerable improvement for increasing numbers of users and local interactions. With regards to \tool's privacy, experiments demonstrate the clustering-based encoding scheme is effective in encoding interactions. In addition, all the experiments relied on a sampling probability~$p=0.5$. This results in a very competitive privacy budget of $\epsilon \approx~0.693$. 

Given that \tool essentially discards half of the data for the sake of privacy, the system performs surprisingly well. In the synthetic benchmarks, \tool traces a similar trend to its non-private counterpart. While for large populations the non-private agents have an advantage, there are more than a few cases where their private counterparts are competitive and even outperform them. Additionally, as the multi-label classification experiments (Figure~\ref{fig:mlc}) indicate, the performance gap between the non-private and differentially-private settings seems to be shrinking to within~\TextMiningDifference and~\MediaMillDifference for the text mining and \MediaMill tasks respectively. \tool's performance on the Criteo dataset is a rare instance where a privacy preserving regime actually improves generalization. The resulting CTR difference of \CriteoDifference in favor or the privacy preserving approach is somewhat surprising. We do provide reasons as to why this is not completely unexpected in our evaluation of results.  

Overall, \tool represents a simple approach to perform privacy-preserving personalization. Our results indicate that \tool becomes a particularly viable option for settings where large user populations participate in a large amount of local interactions, where the performance penalty for privacy is vanishingly small. As future work, we aim to study the behavior of more encoding approaches as well as their interplay with alternative contextual bandit algorithms.


\section*{Acknowledgements}
Mohammad Malekzadeh wishes to thank the Life Sciences Institute at Queen Mary University of London and professor Andrea Caballero for their support through the project. Hamed Haddadi was partially supported by the EPSRC Databox grant (EP/N028260/1).  We would like to thank Dr. Antonio Nappa for his constructive feedback and insights. We would also like to thank anonymous reviewers for their helpful comments and suggestions.
  


\clearpage



\appendix
\section{Artifact Appendix}

\subsection{Abstract}

This section outlines the requisite steps to reproduce the experimental results reported in Section~\ref{sec:eval}. It also provides pointers to  the source code and datasets used for all of the experiments in a GitHub repository.

\subsection{Artifact check-list (meta-information)}
{\small
\begin{itemize}
  \item {\bf Algorithm: } Privacy-Preserving Bandits.
  \item {\bf Program: } Python 3.6 and above.
  \item {\bf Data set: } Synthetic, MediaMill, TextMining, Criteo (see Section \ref{sec:apx_data}).
  \item {\bf Hardware:} We used a 2.3 GHz 8-Core Intel Core i9 with 16 GB RAM. However, any machine that can run Python 3.6 and above is suitable.
  \item {\bf Metrics:} Average Reward, Accuracy, and Click-Through Rate for synthetic, multi-label and online advertising datasets, respectively.
  \item {\bf Output:} Numerical results and their corresponding graphical plots.
  \item {\bf Experiments: } See Section~\ref{sec:apx_exp}. More details are reported in Section~\ref{sec:eval}.
  \item {\bf How much disk space required (approximately)?: } 5GB.
  \item {\bf How much time is needed to prepare workflow (approximately)?: } About 30 minutes to set up the environment for the first time.
  \item {\bf How much time is needed to complete experiments (approximately)?: } About 15 minutes for the small example provided on Synthetic data. To get all results reported for all experiments, one may need about 24 hours.
  \item {\bf Publicly available?: } Yes. 
  \item {\bf Code licenses (if publicly available)?: } MIT license.
  \item {\bf Data licenses (if publicly available)?: } MIT license.
\end{itemize}}

\subsection{Description}

\subsubsection{Availability}

The source code for running all experiments is available in a GitHub repository:

\url{https://github.com/mmalekzadeh/privacy-preserving-bandits} 

The public DOI:

\url{https://doi.org/10.5281/zenodo.3685952}

The artifact evaluated by MLSys 2020 AE chairs:
\url{https://codereef.ai/portal/c/cr-lib/e5a9d2feea4bb3fa}

\subsubsection{Hardware dependencies}
Any hardware that can run Python 3.6 programs and have enough memory to host the required datasets. We used a 2.3 GHz 8-Core Intel Core i9 with 16 GB RAM.
\subsubsection{Software dependencies}
The following python libraries need to be installed: \texttt{iteround}, \texttt{pairing}, \texttt{scikit-multilearn}, \texttt{arff},
\texttt{category\_encoders}, \texttt{matplotlib}, \texttt{tensorflow}, \texttt{keras}. 

\subsubsection{Data sets}\label{sec:apx_data}
The following datasets has been used:
\begin{itemize}
    \item MediaMill~\cite{Snoek2006TheMultimedia}
    \item TextMining dataset~\cite{Srivastava2005DiscoveringSystems}
    \item Criteo dataset from a Kaggle contest\footnote{\url{https://labs.criteo.com/category/dataset}}.
\end{itemize}
By running the experiment, MediaMill and TextMining datasets will be automatically downloaded for the first time. For criteo dataset, in the first time, the script \texttt{create\_datasets.ipynb} should be used. This script is provided in the same repository. 
\subsection{Installation}
A \texttt{requirements.txt} is provided in the repository. After installing \texttt{Python 3.6} or above, you just need to run the following command:
 
\texttt{pip install -r requirements.txt}
 
Then you can continue with the instructions provided in Section~\ref{sec:apx_exp}.

Otherwise, you may need to install and run a \texttt{Jupyter Notebook}\footnote{\url{https://jupyter.org}} then install these libraries via the code provided in the notebooks:
\begin{itemize}
     \item \texttt{$\$$ pip install iteround}
     \item \texttt{$\$$ pip install pairing}
     \item \texttt{$\$$ pip install scikit-multilearn}
     \item \texttt{$\$$ pip install category\_encoders}
     \item \texttt{$\$$ pip install matplotlib}
     \item \texttt{$\$$ pip install tensorflow}
     \item \texttt{$\$$ pip install keras}
\end{itemize}

\subsection{Evaluation and expected result} \label{sec:apx_exp}

The general workflow for reproducing the experimental results reported in Section~\ref{sec:eval} consists of two steps: (i) building an encoder, and subsequently (ii) running the experiments for each dataset. 

All the files necessary to reproduce the results are available as Jupyter Notebooks. Specifically, the GitHub repo contains one Jupyter notebook for building an encoder and a separate Jupyter Notebook explaining step-by-step how to run code and reproduce the results for each of the experiments in Section~\ref{sec:eval}.

\begin{enumerate}
    \item \texttt{build\_an\_encoder.ipynb} Trains a $k$-means clustering~\cite{Sculley2010Web-scaleClustering} based encoder. Depending on the application domain, parameters can be chosen to get  the desired \textit{Encoder} that is consequently used in other experiments. In this file, function
    \texttt{build\_hasher(context\_size, bin\_size, dec\_digits)} gets the required parameters, $d$, $k$, and $q$ explained in Section~\ref{sec:encoding}, to build an \textit{Encoder}.
    
     \item \texttt{a\_synthetic\_exp.ipynb} reproduces the average reward results for synthetic datasets reported in Section \ref{subsec:syn}.
     
     \item \texttt{mlc\_exp.ipynb} reproduces the classification accuracy results for multi-label classification reported in Section \ref{subsec:mlc}.
     
     \item \texttt{criteo\_exp.ipynb} reproduces the click-through rate results for online advertising reported in Section \ref{subsec:oad}.
     
     \item \texttt{a\_synthetic\_figure\_5.ipynb} reproduces the results depicted in Figure~\ref{fig:d_range}.
     
     \item \texttt{to\_plot\_the\_results.ipynb} contains a sample code for plotting results. Note that, one needs to first run a desired experiment, then use this code to plot the results that have been saved during the experiment. 
     
\end{enumerate}

The experiments can be reproduced and even extended by using the following two routines: 
\begin{enumerate}
    \item \texttt{build\_hasher(context\_size, bin\_size, dec\_digits)} gets the required parameters, $d$, $k$, and $q$ explained in Section~\ref{sec:encoding}, to build an \textit{Encoder}.
    \item New experimental setting can be explored by using  \texttt{Simulation.run\_simulation(n\_users, n\_samples, n\_actions, cb\_sampling\_rate, cb\_context\_threshold, neg\_rew\_sam\_rate)} that allow setting the desired set of parameters for each experiment.
     \begin{itemize}
         \item \texttt{n\_users}: number of users $u$.
         \item \texttt{n\_samples}: number of samples per user $t$.
         \item \texttt{n\_actions}: number of actions $|A|$. 
         \item \texttt{cb\_sampling\_rate}: sampling parameter, $p$, for crowd blending (Section~\ref{sec:privacy}).
         \item \texttt{cb\_context\_threshold}: thresholding parameter $l$ (Section~\ref{sec:privacy}).
         \item \texttt{neg\_rew\_sam\_rate}: probability of sharing data if the reward is negative. This is fixed (5\%) across all the experiments reported in this paper.
     \end{itemize}
     The outcome of running this function is Average Reward, Accuracy, or Click-Through Rate for Synthetic, Multi-Label Classification or Online Advertising datasets, respectively. 
\end{enumerate}

\end{document}